\documentclass[twoside,leqno,twocolumn]{article}

\usepackage{times}  
\usepackage{helvet}  
\usepackage{courier}  
\usepackage[hyphens]{url}  
\usepackage{graphicx} 
\usepackage{caption} 

\usepackage[letterpaper]{geometry}

\usepackage{ltexpprt}
\usepackage{hyperref}
\usepackage{subfigure}
\usepackage{amsfonts}
\usepackage{xcolor}
\usepackage{amsmath}
\usepackage{makecell}
\usepackage{multirow}
\usepackage{algorithm}
\usepackage{algorithmic}
\usepackage[symbol]{footmisc}

\usepackage{booktabs}
\usepackage{color, colortbl}

\begin{document}

\newcommand\relatedversion{}
\renewcommand\relatedversion{\thanks{The full version of the paper can be accessed at \protect\url{https://arxiv.org/abs/1902.09310}}} 

\title{Heterogeneous Graph Contrastive Multi-view Learning}
\author{
    Zehong Wang \thanks{School of Mathematics, University of Leeds, Leeds, United Kingdom.}
    \and Qi Li \thanks{Department of Computer Science and Engineering, Shaoxing University, Shaoxing, China}
    \and Donghua Yu \footnotemark[2]
    \and Xiaolong Han \thanks{School of Computer and Software, Nanjing University of Information Science and Technology, Nanjing, China}
    \and Xiao-Zhi Gao \thanks{School of Computing, University of Eastern Finland, Kuopio, Finland.}
    \and Shigen Shen \thanks{School of Information Engineering, Huzhou University, Huzhou, China}  \footnotemark[2]  \thanks{To whom correspondence should be addressed(shigens@zjhu.edu.cn)}
}

\date{}

\maketitle


\fancyfoot[R]{\scriptsize{Copyright \textcopyright\ 2023 by SIAM\\
        Unauthorized reproduction of this article is prohibited}}





\begin{abstract}
    Inspired by the success of Contrastive Learning (CL) in computer vision and natural language processing, Graph Contrastive Learning (GCL) has been developed to learn discriminative node representations on graph datasets. However, the development of GCL on Heterogeneous Information Networks (HINs) is still in the infant stage. For example, it is unclear how to augment the HINs without substantially altering the underlying semantics, and how to design the contrastive objective to fully capture the rich semantics. Moreover, early investigations demonstrate that CL suffers from sampling bias, whereas conventional debiasing techniques are empirically shown to be inadequate for GCL. How to mitigate the sampling bias for heterogeneous GCL is another important problem. To address the aforementioned challenges, we propose a novel Heterogeneous Graph Contrastive Multi-view Learning (HGCML) model. In particular, we use metapaths as the augmentation to generate multiple subgraphs as multi-views, and propose a contrastive objective to maximize the mutual information between any pairs of metapath-induced views. To alleviate the sampling bias, we further propose a positive sampling strategy to explicitly select positives for each node via jointly considering semantic and structural information preserved on each metapath view. Extensive experiments demonstrate HGCML consistently outperforms state-of-the-art baselines on five real-world benchmark datasets. To enhance the reproducibility of our work, we make all the code publicly available at \href{https://github.com/Zehong-Wang/HGCML}{https://github.com/Zehong-Wang/HGCML}. 

\end{abstract}

\section{Introduction}

\begin{figure}
    \includegraphics[width=\linewidth]{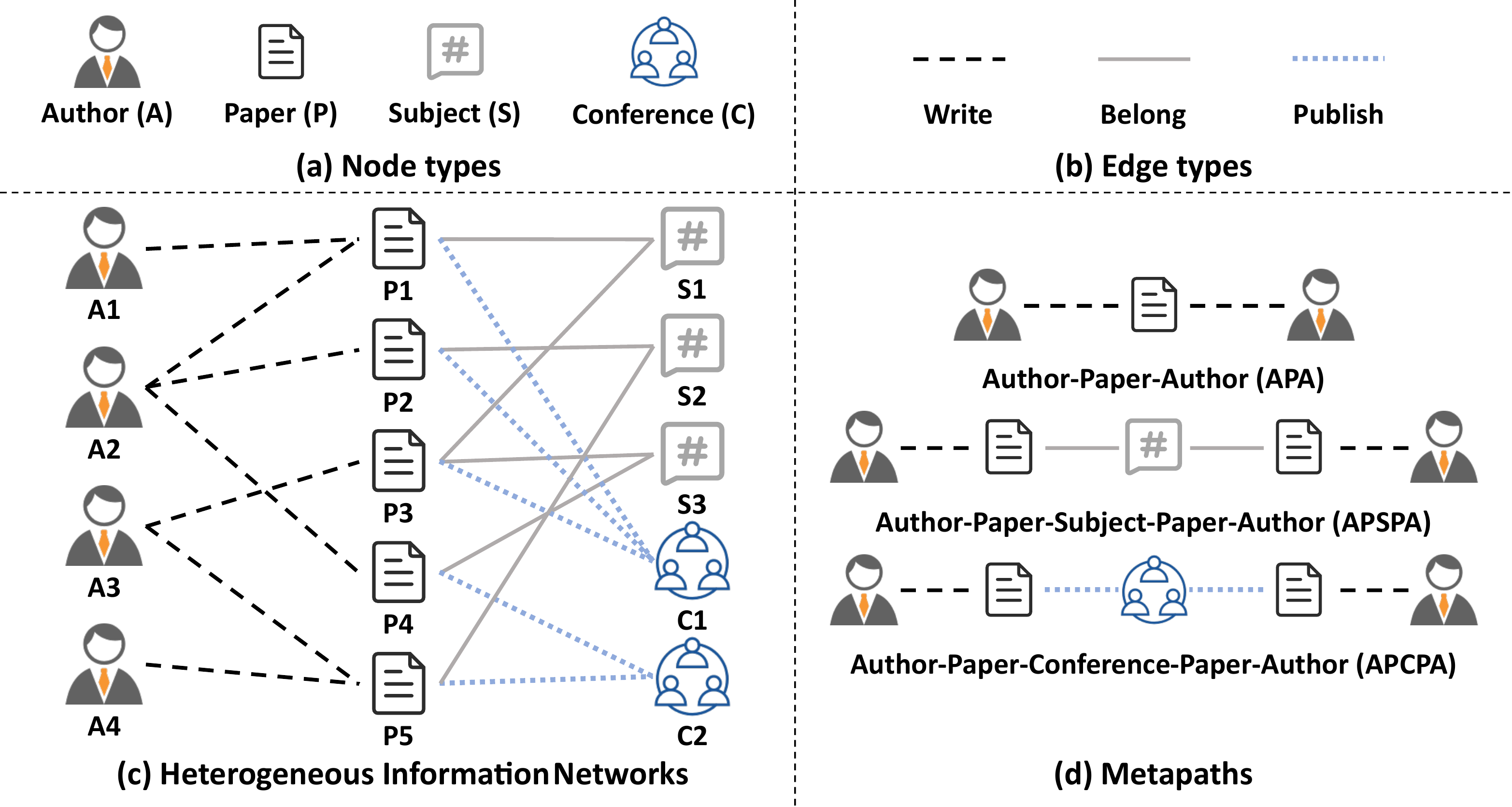}
    \caption{An example of heterogeneous information networks.}
    \label{fig:hin}
\end{figure}

Considering the capacity for modeling complex systems, Heterogeneous Information Networks (HINs) that preserves rich semantics have become a powerful tool for analyzing real-world graphs. As illustrated in Figure \ref{fig:hin}, we present a concise example of a heterogeneous bibliography network with four types of nodes and three types of relations. Recently, Graph Neural Networks (GNNs) \cite{wu_comprehensive_2021} have emerged as a dominant technique in mining graph structure datasets, and its variant, Heterogeneous Graph Neural Networks (HGNNs) \cite{tang_pte_2015,schlichtkrull_modeling_2018,wang_heterogeneous_2019,zhang_heterogeneous_2019,fu_magnn_2020,hu_heterogeneous_2020}, has occupied the mainstream of HIN analysis. In general, HGNNs are trained in an end-to-end manner, which requires abundant, various, and dedicated-designed labels for different downstream tasks. However, in the majority of real-world scenarios, it is highly expensive and/or difficult to collect labels.

Contrastive Learning (CL) \cite{velickovic_deep_2018,he_momentum_2020,chen_simple_2020} that automatically generates supervise signals from data itself is a promising solution for learning representations in a self-supervised manner. By maximizing the confidence (i.e., mutual information) \cite{belghazi_mine_2021} between positive pairs and minimizing the confidence between negative pairs, CL is capable to learn discriminative representations without explicit labels. Inspired by the success of CL in computer vision \cite{he_momentum_2020,chen_simple_2020}, a wide range of Graph Contrastive Learning (GCL) methods have been proposed. For example, DGI \cite{velickovic_deep_2018} exploits a contrast between graph patches (i.e., nodes) and graph summaries, and GRACE \cite{zhu_deep_2020} maximizes the mutual information between the same node in two augmented views. Despite some works generalizing the key idea of CL to homogeneous graphs, there are still three fundamental challenges that need to be addressed in exploring the great potential of CL in heterogeneous graphs:

\textit{\textbf{(1) How to design distinct views?}} Data augmentation that creates corrupted views is shown to be an essential technique to improve the quality of representations \cite{tian_makes_2020}. In GCL, prevalent augmentation methods include edge dropping/adding, node dropping/adding, feature shuffling, and so forth. Although these methods excel in homogeneous graphs \cite{zhu_deep_2020,you_graph_2020}, we believe that they significantly change the latent semantics of HINs. Take a bibliographic network as an example (Figure \ref{fig:hin}); if the link between Author 3 (A3) and Paper 5 (P5) is dropped, the closest path between Author 3 (A3) and Author 4 (A4) will be changed from 2-hop (A3-P5-A4) to 6-hop (A3-P3-S3-P4-C2-P5-A4). To prevent the knowledge perturbation caused by simple augmentation techniques, we propose to leverage metapaths, the composition of semantic relations, to augment datasets. By applying metapaths, we create multiple different yet complementary subgraphs, referred to as metapath views, without altering the underlying semantics while also capturing the high-order relationships on HINs.

\textit{\textbf{(2) How to set proper contrastive objectives?}} The choice of contrastive objectives (i.e., pretext tasks) determines the discriminativeness of representations in downstream tasks. For HIN, the standard choice of pretext tasks is still unclear. Different works present their own solutions. For example, DMGI \cite{park_unsupervised_2020} proposes to use metapaths to learn a shared consensus vector as node representation, HeCo \cite{wang_self-supervised_2021} performs contrast between the aggregation of metapaths (view 1) and network schema (view 2), and HDMI \cite{jing_hdmi_2021} and STENCIL \cite{zhu_structure_2021} iteratively maximize the mutual information between a single metapath and the aggregation of them. Despite these approaches attempting to incorporate the universal knowledge across all metapaths, we think that they actually assume metapaths are independent, which is different from the complementary nature, failing to capture the consistency between metapaths and thus leading to sub-optimality. To directly model the correlation between metapaths, we propose an intuitive yet unexplored contrastive objective that performs contrast between each pair of metapaths. To be specific, the contrast between two augmented views of a metapath (intra-metapath) aims to learn augmentation-invariant representations, and the contrast between two views generated from two sources (inter-metapath) ensures the alignment across metapaths.

\textit{\textbf{(3) How to mitigate the sampling bias?}} Sampling bias indicates that the negative samples, which are randomly selected from the original datasets, are potential to share the same class with the anchor node (i.e., act as false negatives). Empirical, the sampling bias will lead to a significant performance drop. To prevent the issue, existing works \cite{kalantidis_hard_2020,chuang_debiased_2020} aim to select or synthesize hard negatives to mitigate the impact of false negatives. However, these methods are demonstrated to bring limited benefits or even impose adverse impacts on GCL \cite{zhu_empirical_2021,xia_progcl_2022}. To alleviate the issue of false negatives, we propose a positive sampling strategy that collaboratively considers topological and semantic information across metapaths to explicitly decide the positive counterparts for each anchor.

To summarize, we propose a Heterogeneous Graph Contrastive Multi-view Learning (HGCML) model to learn informative node representations on HINs. In particular, we apply metapaths to create multiple views and leverage a GNN model to encode node representations. Then, we employ a novel contrastive objective that aims to maximize the mutual information between any pairs of metapath views (for both intra-metapath and inter-metapath) to explicitly model the complementarity among metapaths, which is neglected in other works. Specifically, we maximize the confidence between two metapaths at node and graph levels to acquire local and global knowledge. To further enhance the expressiveness, we propose a positive sampling strategy that directly picks hard positives for each node based on graph-specific topology and semantics to mitigate the sampling bias inherent in CL. We highlight the contributions as follows:
\begin{itemize}
    \item We propose a heterogeneous graph contrastive multi-view learning framework, named HGCML, to learn discriminative node representations. The model leverages metapaths in HINs to generate multiple views and employs a novel contrastive objective to model the consistency between any pairs of metapath views at node and graph levels.
    \item We propose a positive sampling strategy, which selects the most similar nodes as positive counterparts for each anchor by considering semantics and topology across metapath views, to remedy the sampling bias.
    \item We conduct extensive experiments on five real-world datasets to evaluate the superiority of our model. Experimental results show HGCML outperforms state-of-the-art (SOTA) self-supervised and even supervised baselines.
\end{itemize}

\section{Related Work}

Following the message passing paradigm, GNNs \cite{gilmer_neural_2017,kipf_semi-supervised_2017,hamilton_inductive_2017,velickovic_graph_2018,wang_heterogeneous_2019,fu_magnn_2020,wang2022temporal} have received great attention in recent years for learning representations of nodes in HINs. For example, HAN \cite{wang_heterogeneous_2019} uses attention mechanism to model the correlation between nodes at both metapath-level and semantic-level and MAGNN \cite{fu_magnn_2020} applies metapath encoders to gain fine-grained knowledge preserved in metapaths. To get rid of the impact of metapaths, RGCN \cite{schlichtkrull_modeling_2018} and its variants \cite{zhang_heterogeneous_2019,hu_heterogeneous_2020} directly utilize type-specific matrices to model the relationships between different types of nodes in HINs. Despite these models achieving remarkable performance in mining heterogeneous graph datasets, they fail to be performed in a self-supervised manner.

In another line, GCL that marries the power of GNN and CL has emerged as an important paradigm to learn representations on graphs without annotations. As a pioneering work, DGI \cite{velickovic_deep_2018} treats node embedding and graph summaries as positive pairs and utilizes InfoMAX \cite{belghazi_mine_2021} to optimize the objective. Following this line, MVGRL \cite{hassani_contrastive_2020} proposes to use graph diffusion as an augmentation method to generate multiple views and GraphCL \cite{you_graph_2020} further analyzes the role of augmentations in introducing prior knowledge. Inspired by instance discrimination \cite{wu_unsupervised_2018}, GRACE \cite{zhu_deep_2020} and GCA \cite{zhu_graph_2021} propose to leverage the node-level objective in contrasting to preserve node-level discrimination. In addition, BGRL \cite{thakoor_large_2021} adopts the key idea of BYOL \cite{grill_bootstrap_2020} to perform contrast without negative samples via bootstrapping to save memory consumption.

Meanwhile, some studies have generalized the key idea of GCL on HINs. For instance, HDGI \cite{ren_heterogeneous_2020} extends DGI to heterogeneous graphs and DMGI \cite{park_unsupervised_2020} utilizes a metapath encoder to train consensus vectors as node representations. CKD \cite{wang_collaborative_2022} models the regional and global knowledge between each pair of metapaths, failing to capture node-level properties. CPT-HG \cite{jiang_contrastive_2021} applies relation- and subgraph-level pretext tasks to pre-train HGNN on large-scale HINs, and HDMI \cite{jing_hdmi_2021} introduces a triplet loss to further enhance generalization. However, these methods still do not consider the sampling bias inherent in GCL, inevitably leading to sub-optimality. To mitigate the sampling bias, STENCIL \cite{zhu_structure_2021} and HeCo \cite{wang_self-supervised_2021} propose to apply metapath similarity to measure the hardness between nodes to synthesize hard negatives or select semantic positives. However, these models assume metapaths are independent, and treat the aggregation of metapath-induced subgraphs as a single contrastive view, thus failing to model the consistency and complementarity between metapath views.

Different from the aforementioned methods, our model keeps three distinct advancements: (i) applying metapaths as an augmentation approach to generate multi-views for HINs, instead of treating the aggregation of all metapaths as a single view, which ensures keeping fine-grained and complementary properties for each metapath; (ii) performing contrast between any pairs of metapath-induced subgraphs to learn augmentation-invariant representations for a single metapath and to align the consistency between different metapaths; and (iii) explicitly selecting positive samples for each node via considering topology and semantics preserved on metapath views to mitigate sampling bias.

\section{Preliminary}

\newtheorem{definition}{Definition}

\begin{definition}
    \textbf{Heterogeneous Information Network (HIN)} refers as to a graph consisting of various types of nodes and edges, represented as $\mathcal{G}=\{\mathcal{V}, \mathcal{E}, \mathcal{T}, \mathcal{R}\}$, where $\mathcal{V}$ and $\mathcal{E}$ are the node set and edge set, respectively, and $\mathcal{T}$ and $\mathcal{R}$ denote node types and relation types, associated with a node mapping function $\psi: \mathcal{V} \to \mathcal{T}$ and an edge mapping function $\phi: \mathcal{E} \to \mathcal{R}$. Note that $|\mathcal{T}| + |\mathcal{R}| > 2$.

\end{definition}

\begin{definition}
    \textbf{Metapath.} Metapath $\mathcal{P}_m, m \in \mathcal{M}$, is the composition of relations in HINs, defined as $\mathcal{P}_m := \mathcal{T}_0 \xrightarrow{\mathcal{R}_0} \mathcal{T}_1 \xrightarrow{\mathcal{R}_1}...\xrightarrow{\mathcal{R}_{n}} \mathcal{T}_{n+1}$, where $\mathcal{M}$ is the set of metapaths. For example, we illustrate three metapaths extracted from DBLP in Figure \ref{fig:hin} (d), which describe co-author (APA), co-subject (APSPA), and co-conference (APCPA) relationships.
\end{definition}

\begin{definition}
    \textbf{Metapath-based Neighbors.} Given a metapath $\mathcal{P}_m$, metapath-based neighbors $\mathcal{N}_v^{\mathcal{P}_m}$ is defined as a set of nodes connected to the target node through metapath $\mathcal{P}_m$. For example, in Figure \ref{fig:hin} (c), the metapath-based neighbors of Author 1 via metapath APA is Author 2.
\end{definition}

\begin{figure*}[t]
    \centering
    \includegraphics[width=\linewidth]{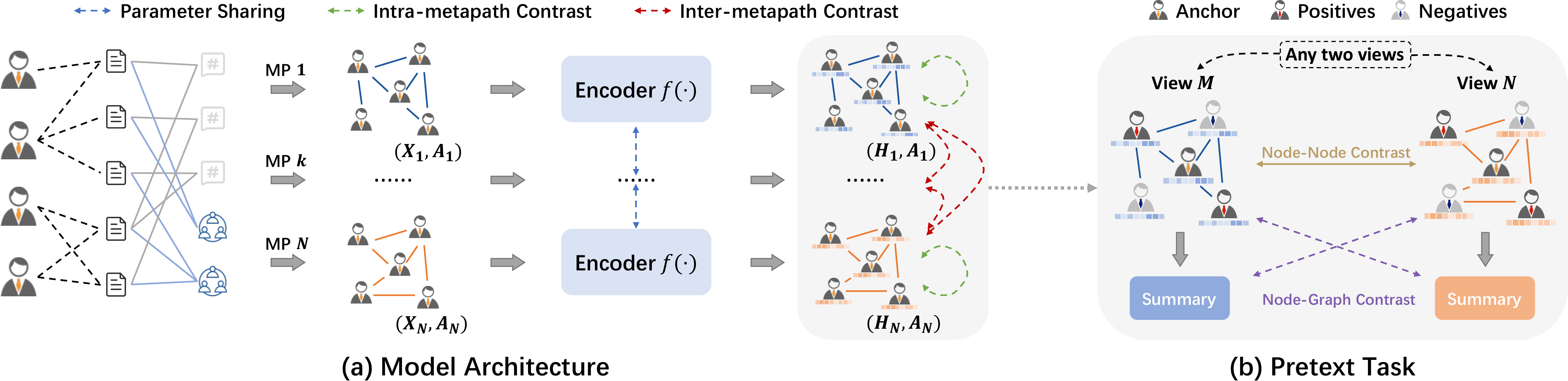}
    \caption{(a) The overview framework of HGCML. Firstly, we generate multi-views via the guide of metapaths (MPs), then leverage a graph neural network (GNN) $f(\cdot)$ to encode the representations. Following, we employ a novel contrastive objective to capture the consistency between each pair of metapath views. The contrast performed between two corrupted versions of a single view is called intra-metapath contrast and the objective applied between two distinct metapath-induced views is referred to as inter-metapath contrast. (b) The proposed pretext task simultaneously learns from graph patches and graph summaries to acquire local and global knowledge. Note that the task is performed between any two metapath-induced views, where $M=N$ denotes intra-metapath contrast and $M \neq N$ indicates inter-metapath contrast. In addition, we perform positive sampling to enhance the expressiveness of node-node contrast.}
    \label{fig:framework}
\end{figure*}

\section{Heterogeneous Graph Contrastive Multi-view Learning}

In this section, we present a heterogeneous graph contrastive multi-view learning framework to learn representations of nodes in HINs. The overview architecture is illustrated in Figure \ref{fig:framework}.

\subsection{Data Augmentation}

For HINs, collectively applying metapaths to construct multi-views is a natural way to supplement the dataset in opposition to simple augmentation techniques. The created multi-views are actually complementary with each other because metapaths depict various facets of the same HIN. Given a set of metapaths $\{\mathcal{P}_0, \mathcal{P}_1, ..., \mathcal{P}_{|\mathcal{M}|}\}$ where $|\mathcal{M}|$ is the number of metapaths, we extract multiple subgraphs (i.e., metapath views) $\{\mathcal{G}_0, \mathcal{G}_1, ..., \mathcal{G}_{|\mathcal{M}|}\}$ from the original graph to sustain the rich semantics preserved in HINs. For the subgraph $\mathcal{G}_m$ generated through metapath $\mathcal{P}_m$, we construct the direct neighborhoods for each node $v$ as its metapath-based neighbors $\mathcal{N}_v^{\mathcal{P}_m}$. Each metapath view is associated with a node feature matrix $\mathbf{X}_m$ and an adjacent matrix $\mathbf{A}_m$. We leverage a GNN encoder $f(\cdot)$ to learn node representation $\{\mathbf{H}_0, ...,\mathbf{H}_m, ..., \mathbf{H}_{|\mathcal{M}|}\}$ from each metapath-induced view, where $\mathbf{H}_m = f(\mathbf{X}_m, \mathbf{A}_m)$. In practice, we leverage additional data augmentations (i.e., feature masking and edge dropping) with specific probabilities $p_f$ and $p_e$ to further corrupt metapath views to make the task to be more difficult, which ensures the learned representations to be more discriminative.

\subsection{Contrastive Objectives}

To distill rich semantics in HINs, we propose a novel contrastive objective to maximize the correlation between any pair of metapath views. In particular, the contrastive objective is collaboratively performed in intra-metapath (i.e., contrast between two corrupted versions of a metapath view) and inter-metapath (i.e, contrast between two views from different metapaths), demonstrated in Figure \ref{fig:framework}(a) with green and red dot lines. We argue that the intra-metapath contrast independently learns the augmentation-invariant latent for each metapath view and the inter-metapath contrast is to align the representations gained from various sources to acquire the complementarity inherent in metapaths. Thus, we thoroughly gain the underlying knowledge maintained in individual metapath views and explicitly model the dependencies between pairs of different metapath views. In addition, the pretext task between two views jointly learns from node- and graph-level knowledge to enhance representativeness, as shown in Figure \ref{fig:framework}(b). Note that in the node-level contrasting, we select hard positives via the proposed sampling strategy to mitigate the sampling bias.

\subsubsection{Node-Node Contrast}

Node-node contrast aims to learn discriminative node representations to boost node-level downstream tasks. Specifically, we perform contrast between the anchor and its positive counterparts in two views to maximize (resp. minimize) the confidence between similar (resp. unassociated) nodes:
\begin{align}
              & \mathcal{L}_{local}^{(m, n)}(u, \mathbb{P}_u) = -log                                                                                                                                                                                                                      \\
    \nonumber & \frac{\sum_{v \in \mathbb{P}_u}\theta(h_u^m, h_v^n)}{\sum\limits_{v \in \mathbb{P}_u}\theta(h_u^m, h_v^n) + \sum\limits_{v \in (\mathcal{V} \setminus \mathbb{P}_u)}\theta(h_u^m, h_v^m) + \sum\limits_{v \in (\mathcal{V} \setminus \mathbb{P}_u)}\theta(h_u^m, h_v^n)},
\end{align}
where the values of $m$ and $n$ can be the same, $h_u^m$ is the representation for node $u$ in view $m$, $\mathbb{P}_u$ denotes the selected positive samples for $u$. We use similarity function $\theta(h_u^m, h_v^n)=e^{\varphi(\rho(h_u^m), \rho(h_v^n)) / \tau}$ to compute the distance between node representations where $\varphi(\cdot, \cdot)$ measures the cosine distance between two vectors, $\rho(\cdot)$ denotes a non-linear projector head that increases the expressiveness, and $\tau$ controls the data distribution. This objective function that pulls semantic similar nodes close and pushes dissimilar nodes away contributes to the discrimination of node representations.

\subsubsection{Node-Graph Contrast}

Different from node-node contrast that learns local semantics across multi-views, we also perform node-graph contrast as an auxiliary task to facilitate the representation learning by injecting metapath-specific knowledge. We define the node-graph contrast objective as follows:
\begin{equation}
    \mathcal{L}_{global}^{(m,n)}(u) = - log(\mathcal{D}(h_u^m, s_m))
    - log(1 - \mathcal{D}(h_u^n, s_m)),
    \label{eq:global loss}
\end{equation}
where the value of $m$ and $n$ can be the same, and $s_m$ is the graph summary of metapath view $\mathcal{G}_m$ calculated via a $READOUT(\cdot)$ function (mean pooling in this paper), and $\mathcal{D}(h, s)=\omega(\rho(h), \rho(s))$ where $\omega(\cdot,\cdot)$ is a discriminator that consists of a bilinear layer $BiLinear(\cdot)$ and a sigmoid function $\sigma(\cdot)$. By imparting global knowledge brought by metapaths, we ensure the representations of nodes are more informative.

\subsubsection{Overall Objective}

The overall objective $\mathcal{J}$ to be maximized is defined as the aggregation of all pairs of metapaths, formally given by
\begin{equation}
    \mathcal{J} = \sum_{m \in \mathcal{M}} \sum_{n \in \mathcal{M}} \sum_{u \in \mathcal{V}} \mathcal{L}_{local}^{(m, n)}(u, \mathbb{P}_u) + \mathcal{L}_{global}^{(m,n)}(u),
\end{equation}
where $\mathcal{M}$ is the set of metapaths. After optimizing the contrastive objective, we perform late fusion function $\eta(\cdot)$ (sum or concatenation) on node representations learned from multiple metapath views to obtain the unified node representations $h_u$ for downstream tasks as
\begin{equation}
    \label{eq:feature fuse}
    h_u = \eta \left( \{h_u^m, m \in \mathcal{M}\} \right),
\end{equation}
where $h_u^m$ denotes the learned representations for node $u$ in metapath-induced view $\mathcal{G}_m$, $\mathcal{M}$ is the metapath set.

\subsection{Positive Sampling Strategy}

\begin{figure}[!t]
    \includegraphics[width=\linewidth]{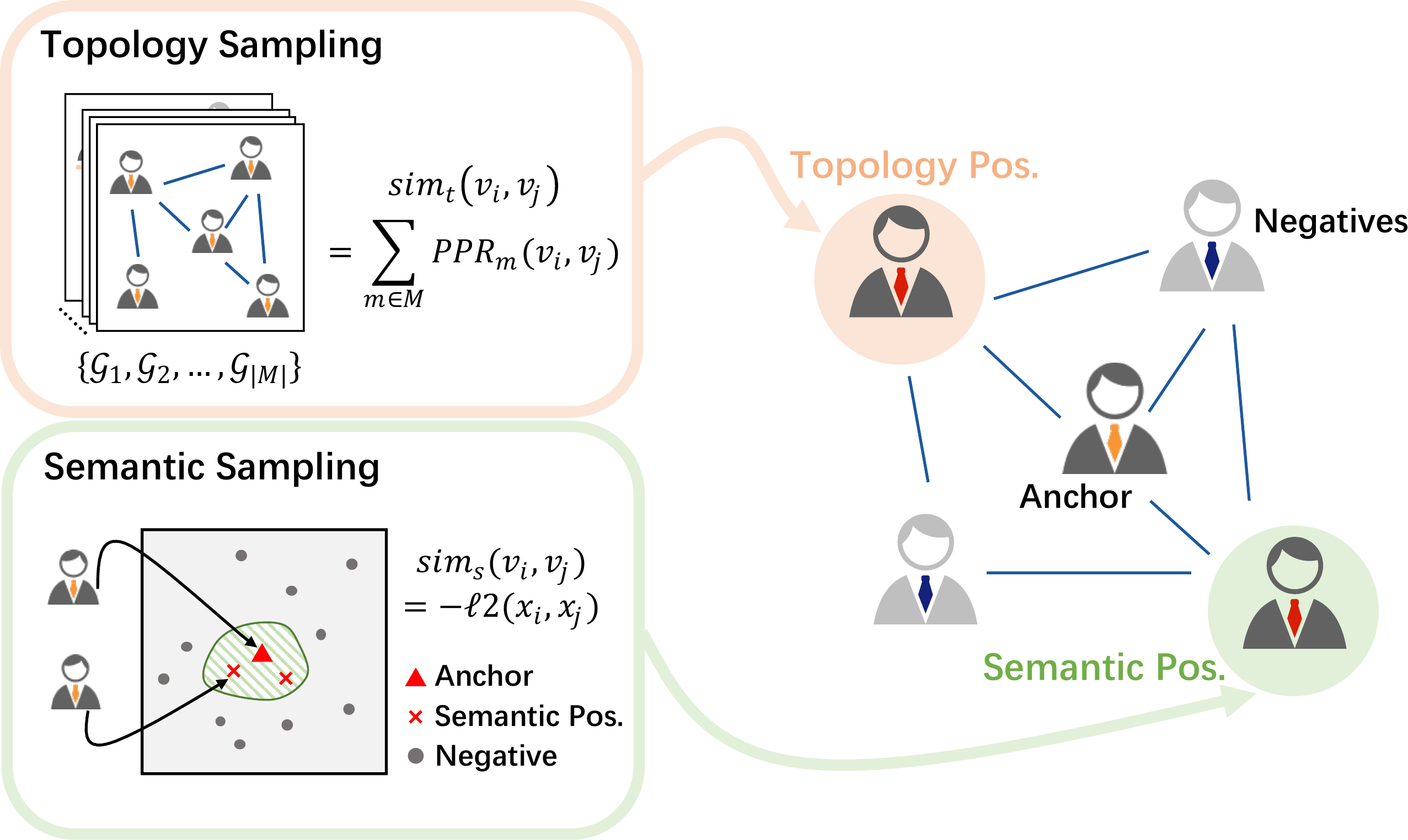}
    \caption{The positive sampling strategy where Personalized PageRank (PPR) is used to measure the topological similarity between nodes, and L2 distance is leveraged to compute the distance between nodes in semantic space to discover semantic associations. }
    \label{fig:sampling}
\end{figure}

Sampling bias is an important problem in CL since false negatives will generate adverse signals. However, existing debiasing techniques \cite{chuang_debiased_2020} are theoretically and empirically verified to lead to severer sampling bias for GCL \cite{xia_progcl_2022}, because the message passing mechanism smooths the node representations. To overcome the deficiency, we propose to leverage two different yet reciprocal similarity measurements (i.e., topology and semantics) to define the distance between nodes, as shown in Figure \ref{fig:sampling}, and explicitly select the most similar nodes as positive samples.

\subsubsection{Topology Positive Sampling}

To analyze the similarity between nodes based on topological structure, we propose to use the graph diffusion kernel \cite{klicpera_diffusion_2019} that assesses the global node importance to compute the distance between two arbitrary nodes. In practice, we apply Personalized PageRank (PPR) score $\mathbf{S}_m$ to measure the node-level relationship for each metapath view $\mathcal{G}_m$, which is defined as
\begin{equation}
    \mathbf{S}_m = \sum_{k=0}^{\infty} \alpha(1-\alpha)^k {(\mathbf{A}_m\mathbf{D}_m^{-1})}^k,
\end{equation}
where $\mathbf{S}_m \in \mathbb{R}^{|\mathcal{V}| \times |\mathcal{V}|}$, $\mathbf{A}_m$, and $\mathbf{D}_m$ are the diffusion matrix, adjacent matrix and diagonal degree matrix for metapath view $\mathcal{G}_m$, respectively, and $\alpha$ denotes teleport probability, whose default is 0.85. Formally, we define the PPR similarity between two nodes $v_i$ and $v_j$ under metapath view $\mathcal{G}_m$ as the $i$-th row and $j$-th column in the diffusion matrix $PPR_m(v_i,v_j) = \mathbf{S}_m[i,j]$. The value in fact describes the stationary probability of starting from $v_i$ to reach $v_j$ via an infinite random walk in the metapath view $m$. Then, we aggregate the PPR scores computed on all metapath-induced views to determine the topological similarity $sim_t(v_i,v_j)$ for each node pair as
\begin{equation}
    sim_t(v_i,v_j) = \sum_{m \in \mathcal{M}} PPR_m(v_i, v_j),
\end{equation}
and select the top-$k$ similar nodes for each anchor as the topology positives $\mathbb{P}^t$.

\subsubsection{Semantic Positive Sampling}

Apart from structural information, graph datasets also preserve rich semantics on the node itself. To measure the semantical similarity between nodes, we propose to utilize a simple metric $sim_s(v_i,v_j)$ to compute the distance between attributes of nodes, which is defined as
\begin{equation}
    sim_s(v_i,v_j) = -l2(x_i, x_j),
\end{equation}
where $x_i$ and $x_j$ denote attributes on nodes $v_i$ and $v_j$, respectively, and $l2(\cdot, \cdot)$ measures the L2-distance between two data points. The attributes for each node will not change across metapath views, thus we only need to process once to calculate the distance between pairs of nodes. Finally, we also select the top-$k$ similar nodes for each anchor as the semantic positives $\mathbb{P}^s$. At the time, we define the positive samples $\mathbb{P}_u$ for node $u$ across metapath views as
\begin{equation}
    \mathbb{P}_u = \mathbb{P}^t_u \cup \mathbb{P}^s_u.
\end{equation}
Note that the positive sampling phase is performed in pre-processing, so the module will not significantly increase the computational complexity.

\section{Experiments}

\definecolor{Gray}{gray}{0.9}

\begin{table*}[!ht]
    \centering
    \resizebox{\textwidth}{!}{
        \begin{tabular}{l|c|ccccc|ccccc}
            \toprule
            \multirow{2}{*}{Methods} & \multirow{2}{*}{Data}                & \multicolumn{5}{c|}{Node Classification} & \multicolumn{5}{c}{Node Clustering}                                                                                                                                                                                                                                                                                                                                                    \\ \cmidrule{3-12}
                                     &                                      & ACM                                      & DBLP                                    & IMDB                                    & Aminer                                  & FreeBase                                & ACM                                     & DBLP                                    & IMDB                                   & Aminer                                  & FreeBase                                \\ \midrule
            DeepWalk                 & $\mathbf{A}$                         & 81.78$\pm$\footnotesize0.04              & 88.09$\pm$\footnotesize0.07             & 56.36$\pm$\footnotesize0.33             & 84.93$\pm$\footnotesize0.09             & 69.63$\pm$\footnotesize0.05             & 41.15$\pm$\footnotesize0.49             & 20.13$\pm$\footnotesize2.57             & 5.97$\pm$\footnotesize0.23             & 30.17$\pm$\footnotesize2.86             & 14.56$\pm$\footnotesize0.08             \\
            MP2vec                   & $\mathbf{A}$                         & 79.82$\pm$\footnotesize0.23              & 87.67$\pm$\footnotesize0.12             & 50.78$\pm$\footnotesize0.18             & 84.14$\pm$\footnotesize0.06             & 69.66$\pm$\footnotesize0.11             & 37.74$\pm$\footnotesize0.09             & 73.77$\pm$\footnotesize0.18             & 2.71$\pm$\footnotesize0.34             & 26.52$\pm$\footnotesize0.36             & 14.93$\pm$\footnotesize1.05             \\
            HIN2vec                  & $\mathbf{A}$                         & 85.23$\pm$\footnotesize0.09              & 91.40$\pm$\footnotesize0.08             & 50.73$\pm$\footnotesize0.23             & 80.77$\pm$\footnotesize0.06             & 67.42$\pm$\footnotesize0.14             & 40.79$\pm$\footnotesize0.49             & 68.83$\pm$\footnotesize1.42             & 3.88$\pm$\footnotesize0.29             & 23.76$\pm$\footnotesize0.64             & 14.26$\pm$\footnotesize2.03             \\
            HERec                    & $\mathbf{A}$                         & 67.15$\pm$\footnotesize0.85              & 90.75$\pm$\footnotesize0.39             & 49.12$\pm$\footnotesize0.22             & 80.63$\pm$\footnotesize0.10             & 68.04$\pm$\footnotesize0.19             & 45.39$\pm$\footnotesize2.11             & 70.38$\pm$\footnotesize3.29             & 4.39$\pm$\footnotesize1.01             & 31.05$\pm$\footnotesize0.69             & 15.32$\pm$\footnotesize1.05             \\ \midrule
            DGI                      & $\mathbf{X}, \mathbf{A}$             & 88.44$\pm$\footnotesize0.30              & 90.16$\pm$\footnotesize0.60             & 52.00$\pm$\footnotesize0.94             & 83.24$\pm$\footnotesize0.19             & 68.42$\pm$\footnotesize0.31             & 43.47$\pm$\footnotesize2.25             & 54.44$\pm$\footnotesize2.07             & 4.09$\pm$\footnotesize1.88             & 29.80$\pm$\footnotesize1.86             & 15.16$\pm$\footnotesize1.13             \\
            GRACE                    & $\mathbf{X}, \mathbf{A}$             & 87.64$\pm$\footnotesize0.31              & 91.28$\pm$\footnotesize0.07             & 54.80$\pm$\footnotesize0.82             & 83.43$\pm$\footnotesize0.22             & 69.25$\pm$\footnotesize0.14             & 46.50$\pm$\footnotesize4.58             & 67.98$\pm$\footnotesize1.32             & 1.58$\pm$\footnotesize1.12             & 24.12$\pm$\footnotesize5.24             & 16.23$\pm$\footnotesize3.37             \\
            DMGI                     & $\mathbf{X}, \mathbf{A}$             & 76.76$\pm$\footnotesize1.23              & 91.60$\pm$\footnotesize0.66             & 51.16$\pm$\footnotesize0.63             & 79.19$\pm$\footnotesize0.32             & 67.69$\pm$\footnotesize0.21             & 52.53$\pm$\footnotesize1.73             & 67.41$\pm$\footnotesize0.07             & 5.45$\pm$\footnotesize0.12             & 28.32$\pm$\footnotesize0.44             & 12.35$\pm$\footnotesize0.35             \\
            STENCIL                  & $\mathbf{X}, \mathbf{A}$             & 88.23$\pm$\footnotesize0.91              & 92.56$\pm$\footnotesize0.22             & 57.83$\pm$\footnotesize0.62             & 84.61$\pm$\footnotesize0.53             & 68.26$\pm$\footnotesize0.21             & 56.67$\pm$\footnotesize2.51             & 71.40$\pm$\footnotesize1.93             & 8.25$\pm$\footnotesize1.09             & 29.99$\pm$\footnotesize2.69             & 13.19$\pm$\footnotesize1.10             \\
            HeCo                     & $\mathbf{X}, \mathbf{A}$             & 88.97$\pm$\footnotesize1.12              & 92.24$\pm$\footnotesize0.48             & 52.12$\pm$\footnotesize0.72             & 85.22$\pm$\footnotesize0.10             & 69.02$\pm$\footnotesize0.07             & 56.93$\pm$\footnotesize1.59             & 70.03$\pm$\footnotesize1.25             & 7.41$\pm$\footnotesize1.26             & 30.61$\pm$\footnotesize3.81             & 12.07$\pm$\footnotesize1.47             \\
            \rowcolor{Gray}
            \textbf{HGCML}           & $\mathbf{X}, \mathbf{A}$             & 91.02$\pm$\footnotesize0.13              & 93.29$\pm$\footnotesize0.12             & 60.75$\pm$\footnotesize0.71             & 86.63$\pm$\footnotesize0.11             & 71.41$\pm$\footnotesize0.04             & 65.13$\pm$\footnotesize1.33             & 73.28$\pm$\footnotesize0.76             & \textbf{9.34$\pm$\footnotesize0.86}    & \textbf{36.10$\pm$\footnotesize2.44}    & 15.46$\pm$\footnotesize1.65             \\
            \rowcolor{Gray}
            \textbf{HGCML-P}         & $\mathbf{X}, \mathbf{A}$             & \textbf{91.34$\pm$\footnotesize0.17}     & \textbf{93.44$\pm$\footnotesize0.08}    & \textbf{61.02$\pm$\footnotesize0.49}    & \textbf{87.03$\pm$\footnotesize0.06}    & \textbf{71.53$\pm$\footnotesize0.14}    & \textbf{65.75$\pm$\footnotesize1.62}    & \textbf{74.53$\pm$\footnotesize0.48}    & 8.95$\pm$\footnotesize1.06             & 35.62$\pm$\footnotesize1.74             & \textbf{16.26$\pm$\footnotesize2.56}    \\ \midrule
            GCN                      & $\mathbf{X}, \mathbf{A}, \mathbf{Y}$ & \underline{89.87$\pm$\footnotesize0.79}  & 92.04$\pm$\footnotesize1.03             & \underline{58.42$\pm$\footnotesize1.42} & 85.42$\pm$\footnotesize0.48             & 69.13$\pm$\footnotesize2.51             & 58.14$\pm$\footnotesize0.90             & 77.71$\pm$\footnotesize1.35             & \underline{8.59$\pm$\footnotesize0.84} & \underline{37.80$\pm$\footnotesize1.69} & 15.77$\pm$\footnotesize2.97             \\
            GAT                      & $\mathbf{X}, \mathbf{A}, \mathbf{Y}$ & 88.84$\pm$\footnotesize0.61              & 92.51$\pm$\footnotesize1.28             & 57.97$\pm$\footnotesize1.64             & 84.37$\pm$\footnotesize0.42             & 70.42$\pm$\footnotesize0.55             & \underline{62.22$\pm$\footnotesize3.67} & 72.06$\pm$\footnotesize1.61             & 8.04$\pm$\footnotesize1.76             & 36.81$\pm$\footnotesize0.66             & 15.44$\pm$\footnotesize1.32             \\
            HAN                      & $\mathbf{X}, \mathbf{A}, \mathbf{Y}$ & 89.50$\pm$\footnotesize1.21              & \underline{93.27$\pm$\footnotesize0.58} & 54.78$\pm$\footnotesize1.01             & \underline{85.90$\pm$\footnotesize0.43} & \underline{70.98$\pm$\footnotesize1.07} & 60.98$\pm$\footnotesize2.38             & \underline{78.20$\pm$\footnotesize0.83} & 6.80$\pm$\footnotesize2.32             & 35.37$\pm$\footnotesize0.48             & \underline{16.38$\pm$\footnotesize1.73} \\ \bottomrule
        \end{tabular}
    }
    \caption{Performance of node classification and clustering on five benchmark datasets in terms of micro-F1 and normalized mutual information (NMI). Boldfaces and underlines denote the best performance among self-supervised and supervised methods, respectively. For our model, we use the suffix -P to indicate the positive sampling version.}
    \label{tab:node classification clustering}
\end{table*}

\subsection{Experimental Setup}

\subsubsection{Datasets and Baselines} To demonstrate the superiority of HGCML over SOTA, we conduct extensive experiments on five public benchmark datasets, including ACM, DBLP, IMDB, Aminer, and FreeBase. We evaluate the performance of our model against various baselines from shallow graph representation learning algorithms, including DeepWalk \cite{perozzi_deepwalk_2014}, Metapath2vec(MP2vec) \cite{dong_metapath2vec_2017}, HIN2vec \cite{fu_hin2vec_2017}, HERec \cite{shi_heterogeneous_2019}, to GCL methods (e.g., DGI \cite{velickovic_deep_2018}, GRACE \cite{zhu_deep_2020}, DMGI \cite{park_unsupervised_2020}, STENCIL \cite{zhu_structure_2021}, HeCo \cite{wang_self-supervised_2021}) to supervised GNNs, like GCN \cite{kipf_semi-supervised_2017}, GAT \cite{velickovic_graph_2018}, HAN \cite{wang_heterogeneous_2019}. Note that DMGI, STENCIL, and HeCo are dedicated for heterogeneous graphs.

\subsubsection{Evaluation Protocol}

We evaluate HGCML on node classification and node clustering. For node classification, we use Micro-F1 as the metric and follow the linear protocol that utilizes the learned graph encoder as a feature extractor to train a simple linear classifier with 20\% random samples as the training set. For node clustering, we apply $K$-means to generate clusters and utilize Normalized Mutual Information (NMI) as the metric. To mitigate the impact of initialized centroids, we perform 10 times clustering and report the average results. For all baselines, we run 10 times and present the average scores with standard deviations. For DGI, GRACE, GCN, and GAT, we create homogeneous graphs based on metapaths and report the best results.

\begin{table}[!t]
    \centering
    \resizebox{\linewidth}{!}{
        \begin{tabular}{cc|cc|ccccc}
            \toprule
            Intra-  & Inter-  & Local   & Global  & ACM            & DBLP           & IMDB           & Aminer         & FreeBase       \\ \midrule
            $\surd$ & -       & $\surd$ & -       & 89.20          & 91.94          & 58.71          & 84.95          & 69.61          \\
            $\surd$ & -       & -       & $\surd$ & 83.20          & 90.95          & 48.56          & 83.42          & 69.38          \\
            $\surd$ & -       & $\surd$ & $\surd$ & 90.52          & 92.52          & 60.12          & 86.17          & 71.16          \\
            -       & $\surd$ & $\surd$ & $\surd$ & 88.32          & 92.44          & 59.80          & 85.92          & 70.65          \\
            \rowcolor{Gray}
            $\surd$ & $\surd$ & $\surd$ & $\surd$ & \textbf{91.02} & \textbf{93.29} & \textbf{60.75} & \textbf{86.63} & \textbf{71.41} \\ \bottomrule
        \end{tabular}
    }
    \caption{Ablation study of the proposed HGCML for pretext tasks on node classification, where the intra- and inter- are abbreviations of intra-metapath and inter-metapath contrasts, and local and global indicate node-node and node-graph contrasts, respectively.}
    \label{tab:ablation contrast}
\end{table}

\subsubsection{Implementation Details}

We leverage a 1-layer GCN as the encoder for each metapath-induced view. The parameters are initialized via Xavier initialization and we apply Adam as the optimizer. We perform grid search to tune the learning rate from $5e-4$ to $5e-3$, the value of temperature from $0.2$ to $0.8$, the corrupt rate from $0.1$ to $0.7$, and the number of positives from $0$ to $128$. Moreover, we set early stop to $20$ epochs, node dimension to $64$, activation function to $ReLU(\cdot) = max(\cdot, 0)$, and use concatenation as the fusion function in ACM and DBLP, and summation in other datasets.

\begin{table}[!t]
    \centering
    \resizebox{\linewidth}{!}{
        \begin{tabular}{cc|ccccc}
            \toprule
            Topology Pos. & Semantic Pos. & ACM            & DBLP           & IMDB           & Aminer         & FreeBase       \\ \midrule
            -             & -             & 91.02          & 93.29          & 60.75          & 86.63          & 71.41          \\
            $\surd$       & -             & 91.17          & 93.34          & 59.92          & 86.88          & 71.43          \\
            -             & $\surd$       & 90.88          & 93.35          & 58.28          & 86.91          & 71.45          \\
            \rowcolor{Gray}
            $\surd$       & $\surd$       & \textbf{91.34} & \textbf{93.44} & \textbf{61.02} & \textbf{87.03} & \textbf{71.53} \\ \bottomrule
        \end{tabular}}
    \caption{Ablation study of the proposed HGCML for positive sampling strategies on node classification with four variants; $\surd$ denotes the specific type of positives are selected. }
    \label{tab:ablation pos}
\end{table}

\begin{figure*}[!t]
    \centering
    \subfigure[ACM]{
        \includegraphics[width=0.2\linewidth]{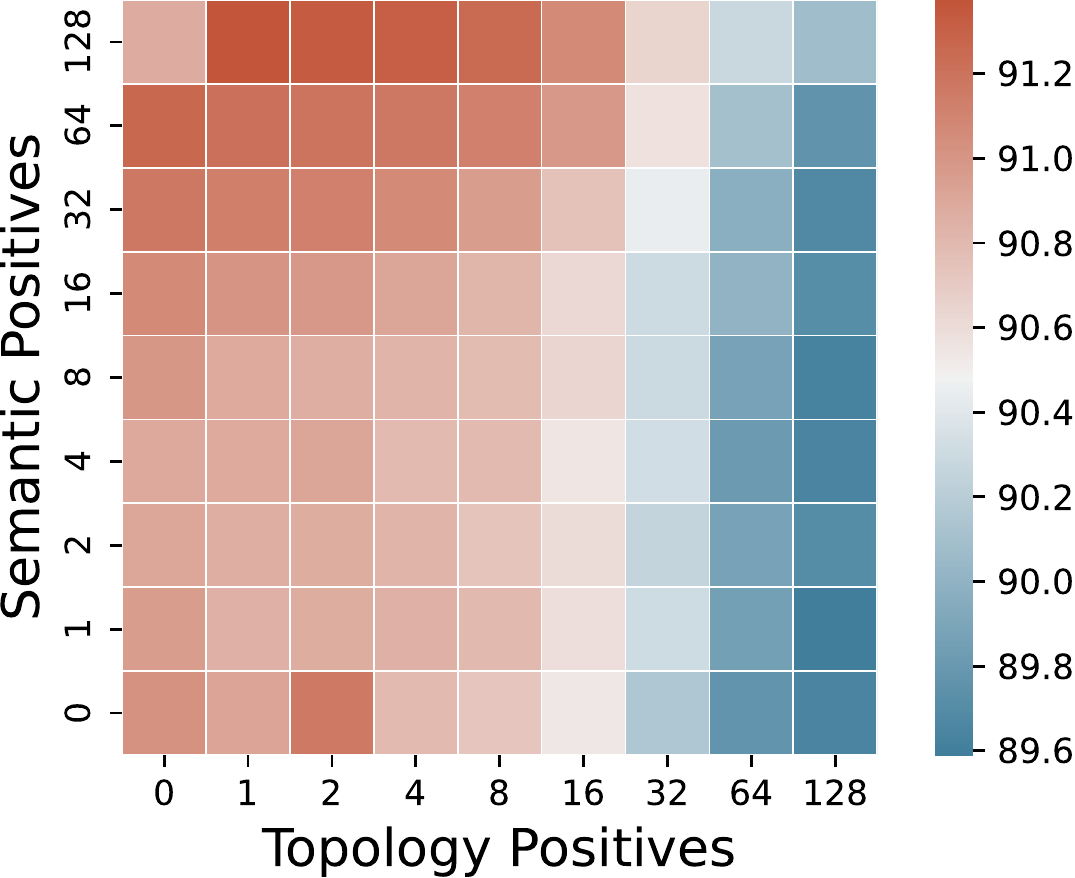}
        \label{fig:pos acm}
    }
    \subfigure[DBLP]{
        \includegraphics[width=0.2\linewidth]{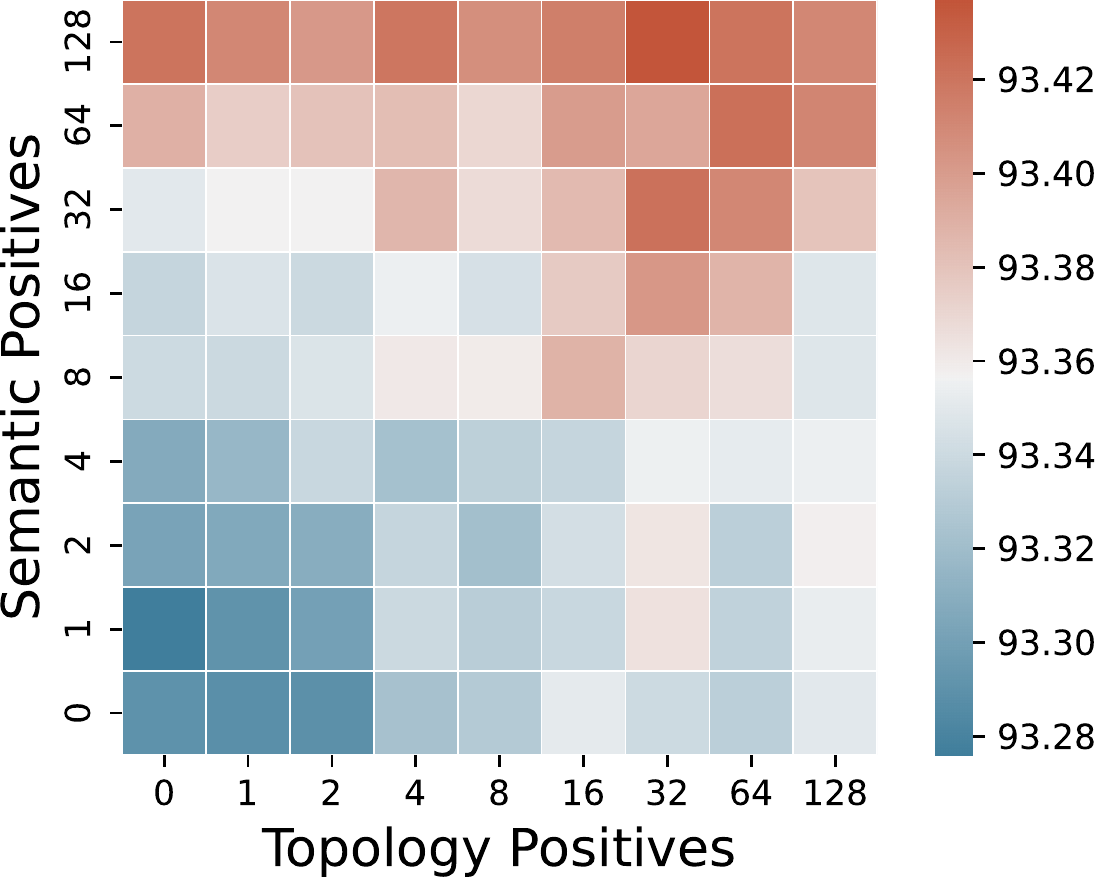}
        \label{fig:pos dblp}
    }
    \subfigure[AMiner]{
        \includegraphics[width=0.2\linewidth]{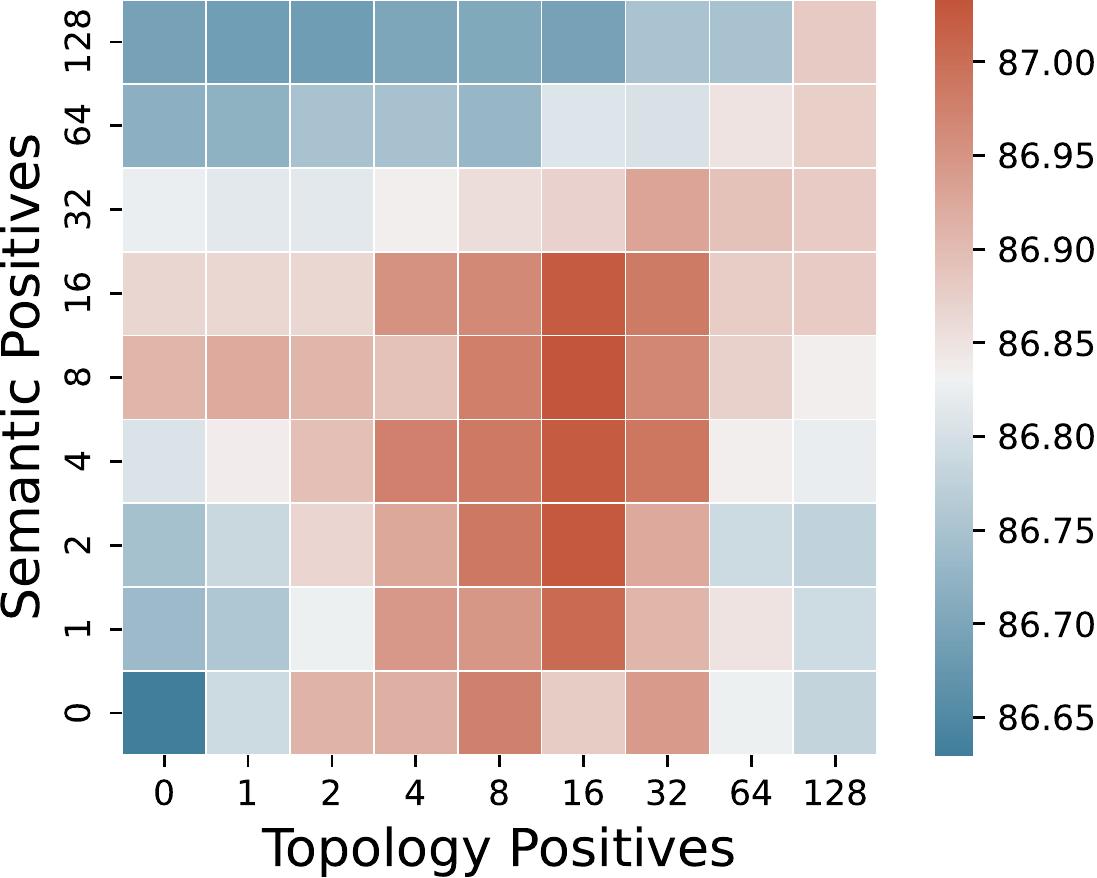}
        \label{fig:pos aminer}
    }
    \subfigure[IMDB]{
        \includegraphics[width=0.2\linewidth]{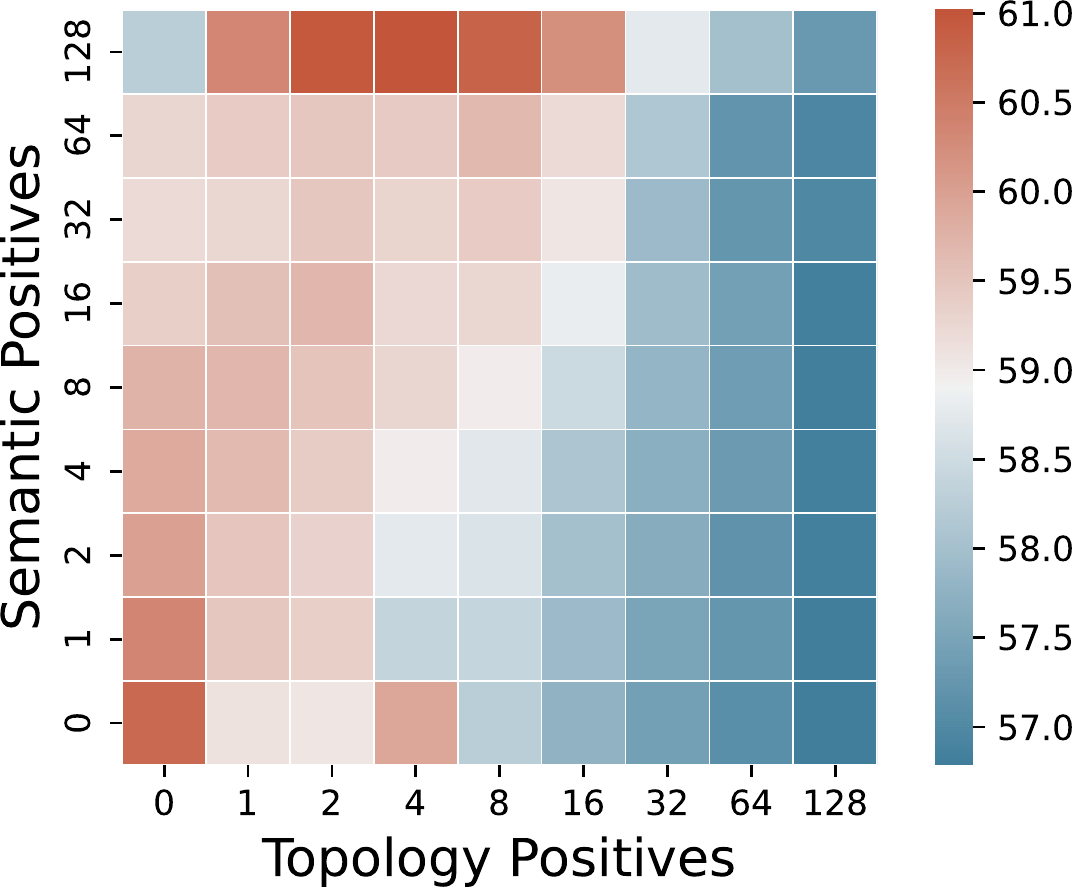}
        \label{fig:pos imdb}
    }
    \caption{Hyperparameter sensitivity of positive sampling thresholds on node classification. Note that when no extra positives are selected (i.e., the left bottom corner), the model picks the anchor node itself as the positive sample. }
    \label{fig:pos analysis}
\end{figure*}

\begin{figure*}[!t]
    \centering
    \subfigure[ACM]{
        \centering
        \includegraphics[width=0.2\linewidth]{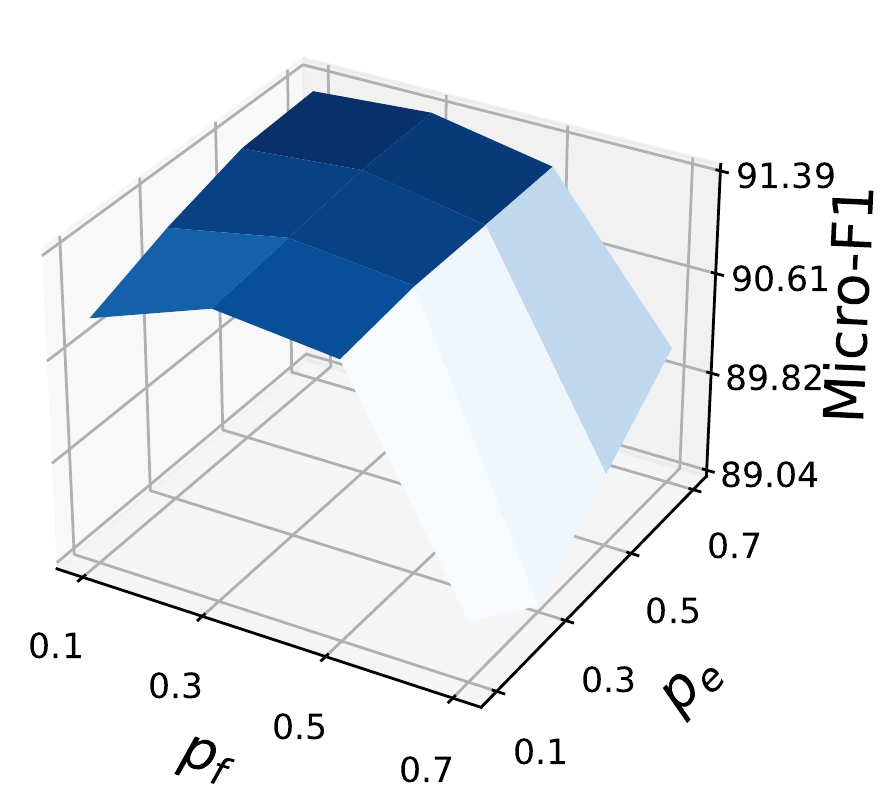}
        \label{fig:corrupt acm}
    }
    \subfigure[DBLP]{
        \centering
        \includegraphics[width=0.2\linewidth]{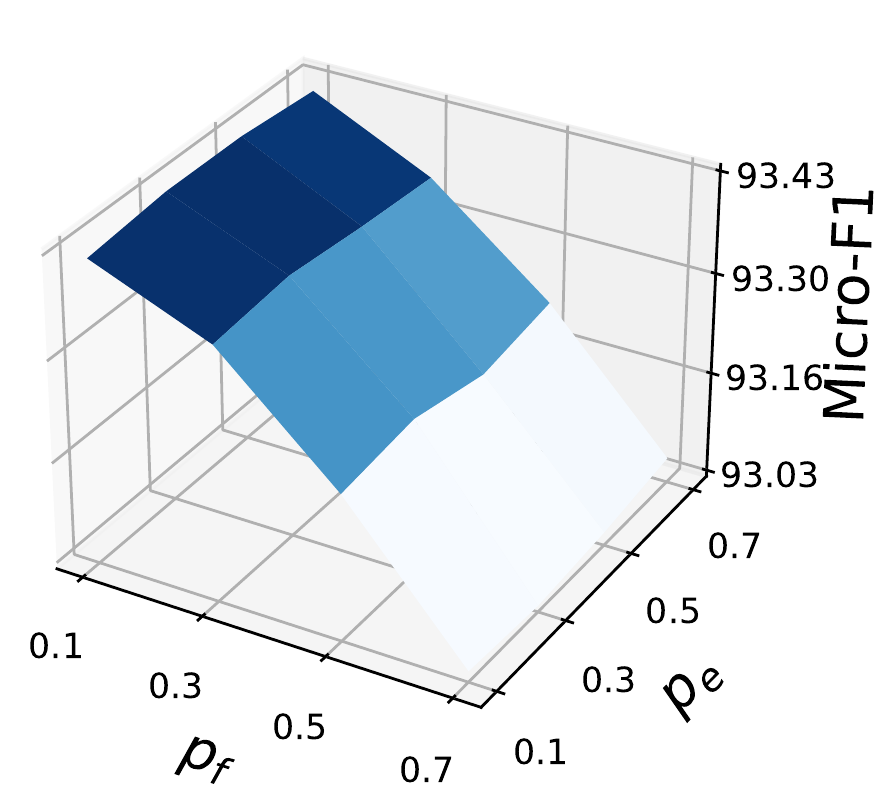}
        \label{fig:corrupt dblp}
    }
    \subfigure[AMiner]{
        \centering
        \includegraphics[width=0.2\linewidth]{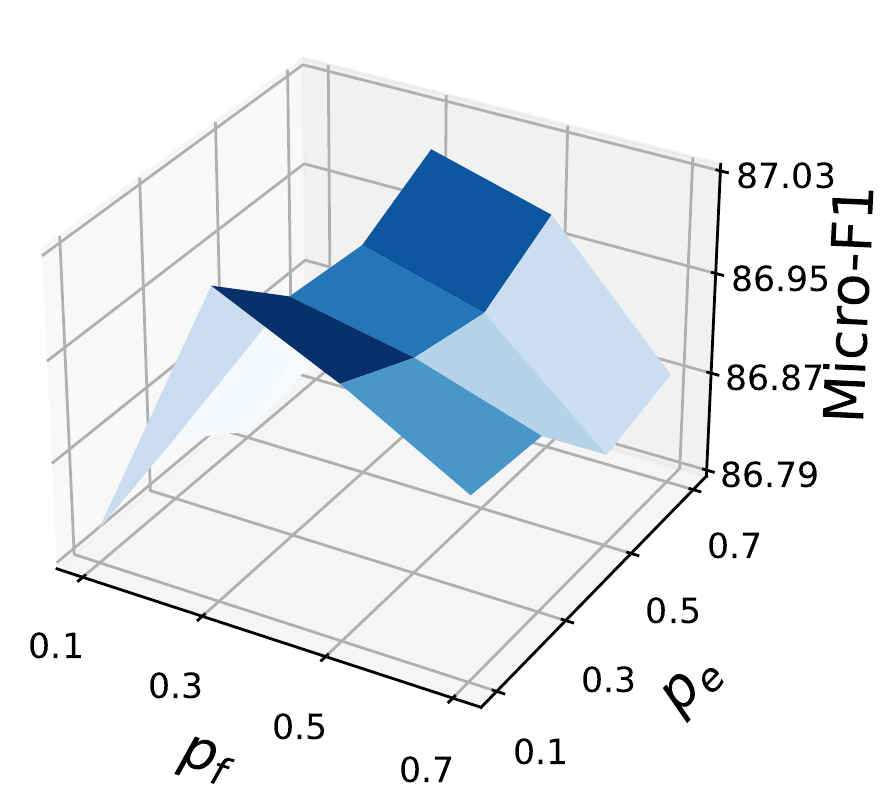}
        \label{fig:corrupt aminer}
    }
    \subfigure[FreeBase]{
        \centering
        \includegraphics[width=0.2\linewidth]{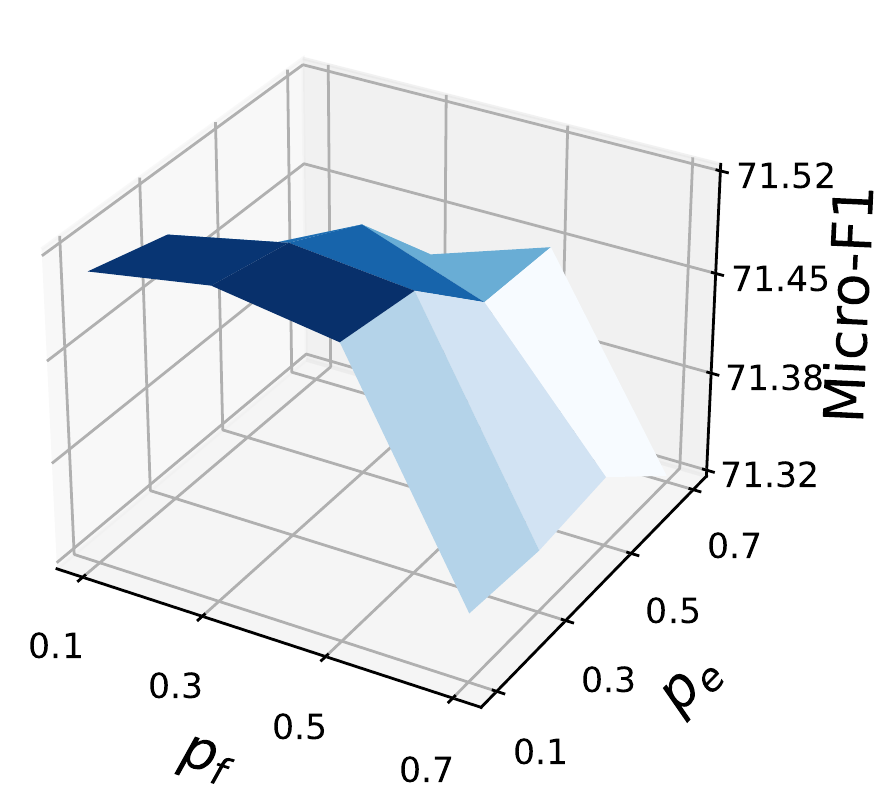}
        \label{fig:corrupt freebase}
    }
    \caption{Hyperparameter sensitivity of augmentations (edge dropping $p_e$ and feature masking $p_f$) on node classification.}
    \label{fig:corrupt analysis}
\end{figure*}

\subsection{Quantitative Results}

We report the quantitative result of node classification and node clustering with standard deviations in Table \ref{tab:node classification clustering}. From the table, we observe that GCL methods generally perform better than shallow unsupervised baselines, since the instance discrimination applied on CL captures underlying semantics preserved in HINs but the graph reconstruction adopted in classical methods only considers the topological structure. Our models (HGCML and HGCML-P) consistently outperform SOTA self-supervised graph learning methods across all datasets by a large margin on supervised classification and unsupervised clustering tasks, and even achieve competitive results compared to supervised baselines. Beyond that, the performance of HGCML-P (positive sampling version) is commonly better than its vanilla version that performs contrast between the same node in different views, demonstrating the necessity of introducing correlated nodes as positives to mitigate the sampling bias. Compared with heterogeneous graph contrastive learning methods (i.e., DMGI, STENCIL, and HeCo), our model always acquires higher scores in both classification and clustering. We assume that it is because (1) the intra-metapath contrast is performed between nodes on two corrupted views induced from the same metapath to learn the discriminative representations and (2) the inter-metapath contrast captures the complementarity between metapaths instead of treating the aggregation of them as a single view under the independent assumption applied in mentioned baselines.

\subsection{Ablation Study}

\subsubsection{Pretext Task}

To verify the role of each component in the contrastive objective, we perform ablation studies, as shown in Table \ref{tab:ablation contrast}, to compare the performance of multiple variants on node classification. From the table, we observe that (1) the node-node contrast provides better discrimination ability compared with the node-graph contrast since the fine-grained information (patches) is leveraged in learning representations. When the node- and graph-level objectives are jointly optimized, the performance is significantly improved, showing the necessity of simultaneously modeling local- and global-level dependencies. (2) The intra-metapath contrast is essential in promoting the learning procedure, reflected in the competitive performance obtained in the initialized variant (Intra- \& Node) against SOTA self-supervised baselines. (3) The variant with full components persistently achieves the best performance since the intra-metapath contrast captures the latent semantics of each metapath-induced view and the inter-metapath contrast aligns the consistency between metapaths. If one of them is removed, we cannot thoroughly model the relationship between metapaths, thus encountering model degradation.

\subsubsection{Positive sampling strategy}

We also conduct experiments to evaluate the impact of positive sampling, as presented in Table \ref{tab:ablation pos}. We can find that the selected positives indeed improve the performance by implicitly defining hard negatives. In addition, the significance of these two positive sampling strategies depends on the choice of datasets, i.e, there is no obvious superiority between topology positives and semantic positives. However, when they are jointly leveraged, our model achieves the best scores. The phenomenon demonstrates that the selected positives based on different strategies are distinct yet complementary.

\subsection{Hyperparameter Analysis}

\subsubsection{Positive Sampling Thresholds}

In the above section, we analyze the impact of positive sampling strategies in enhancing the quality of representations, here we delve into the positive sampling thresholds to provide a further examination, illustrated in Figure \ref{fig:pos analysis}. As we can see, the best performance is achieved with a large number of semantic positives and a small number of topology positives (ACM, DBLP, IMDB). When the number of topology positives is too large, the performance generally encounters a drop. We assume the phenomena derive from the inherent property of defined similarity functions. To be specific, the semantical similarity is independently measured on attributes of nodes in the representation space, whereas the topological similarity is calculated based on the adjacent matrix, which makes the function naturally biased to nodes with dense connections. Thus, when the number of topology positives is too large, there will contain too many noisy nodes. The Aminer does not follow the observation on the other datasets, whose best performance is achieved when the number of semantic and topology positives are both small. We consider it is because the attributes on Aminer are generated by DeepWalk, a random walk-based algorithm that is biased to hub nodes in the learning procedure.

\subsubsection{Augmentation Probabilities}

In this section, we present the impact of two critical data augmentation hyperparameters, i.e., edge dropping $p_e$ and feature masking $p_f$, in Figure \ref{fig:corrupt analysis}. We have the following observations. (1) A relatively low corrupt probability 0.3 is desirable to achieve competitive results. (2) When the dropping probability is too large, we face a model degradation because the semantics and/or structures for each metapath-induced view are significantly corrupted, failing to preserve enough augmentation-invariant information. (3) Despite the performance dramatically fluctuating under different probability combinations, the absolute value between the maximum and minimum is generally less than 0.5 except ACM, showing the robustness of HGCML on augmentation probabilities.

\begin{figure}[!t]
    \centering
    \subfigure[ACM]{
        \centering
        \includegraphics[width=0.45\linewidth]{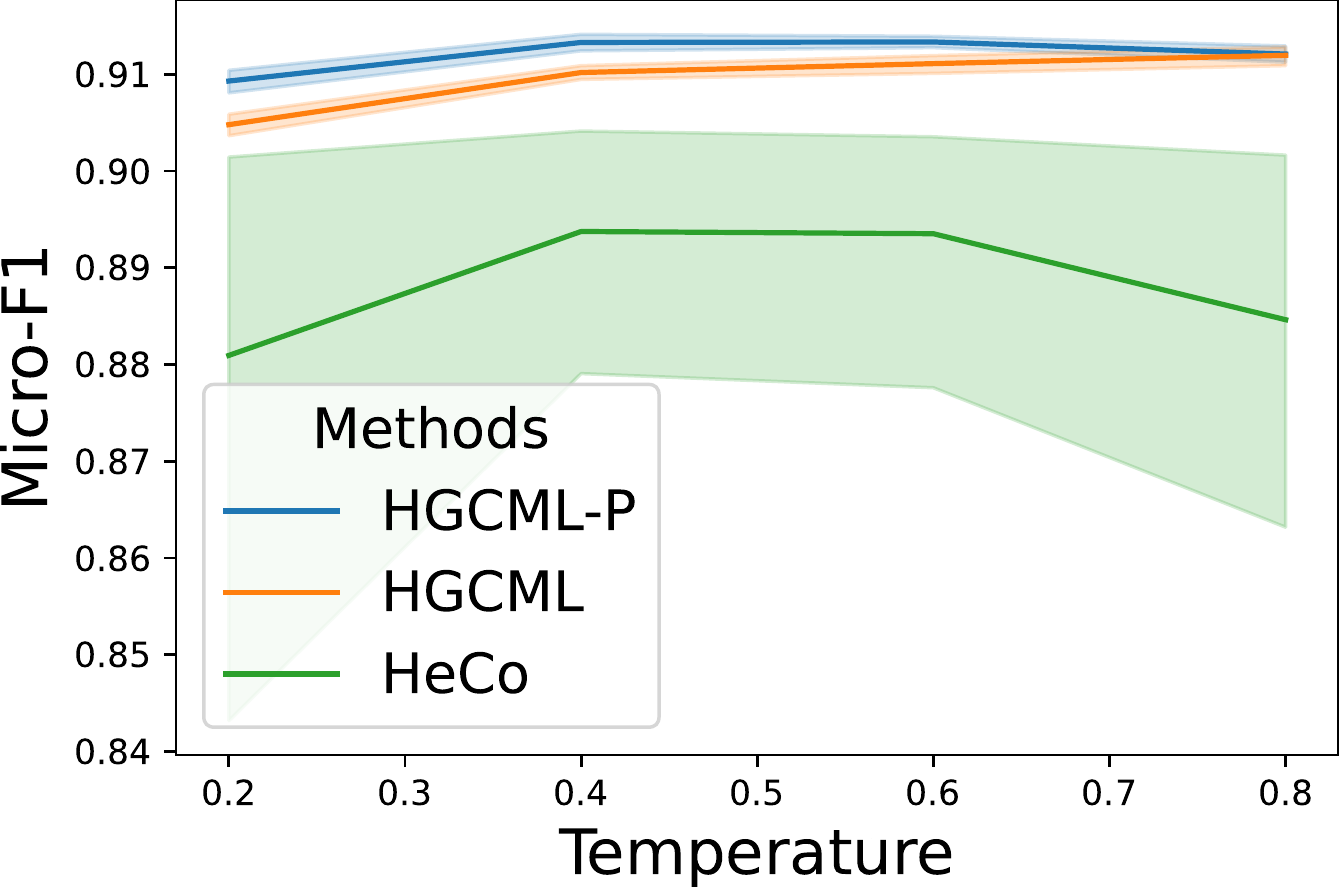}
        \label{fig:tau acm}
    }
    \subfigure[DBLP]{
        \centering
        \includegraphics[width=0.45\linewidth]{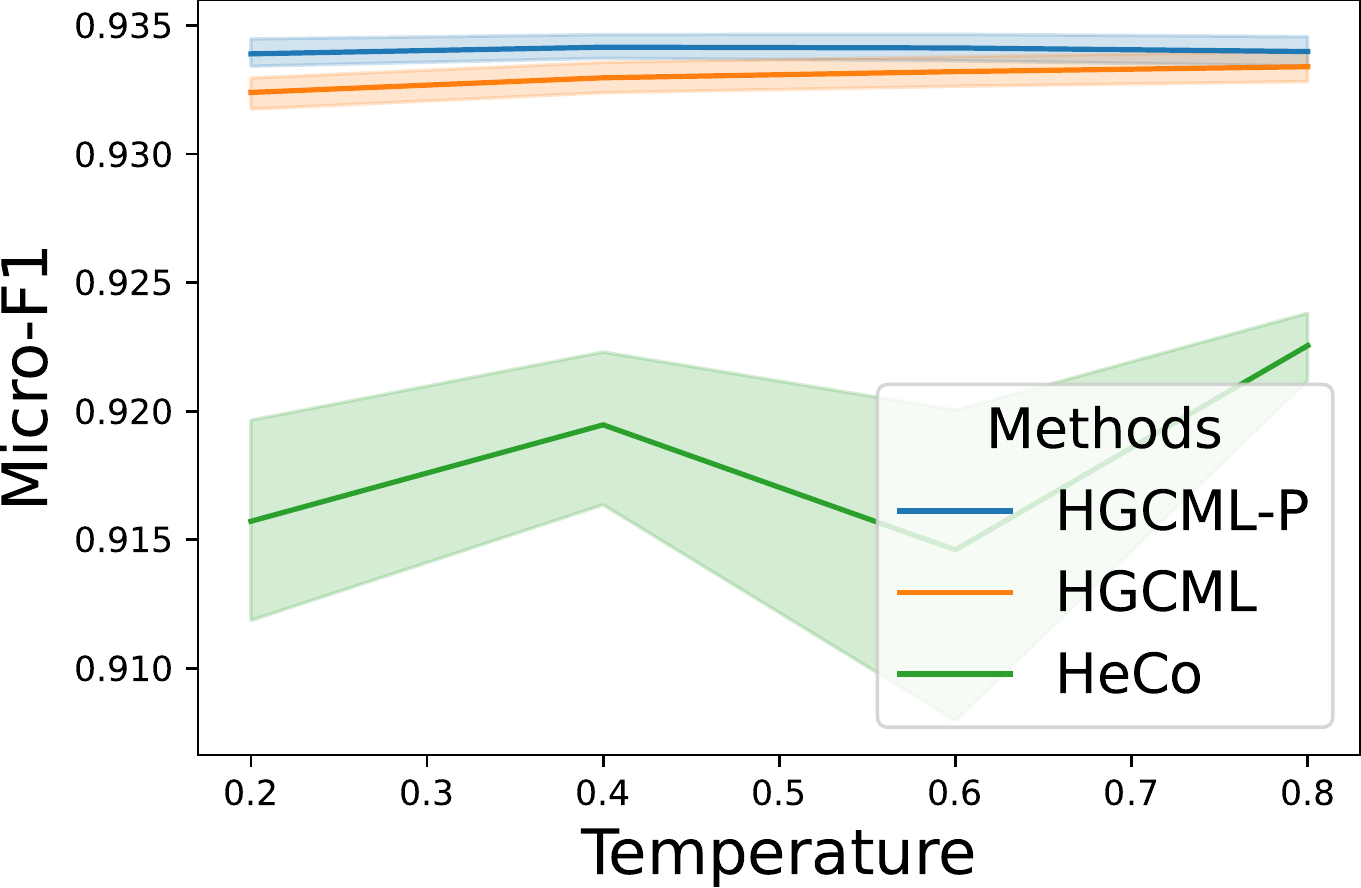}
        \label{fig:tau dblp}
    }
    \caption{Hyperparameter sensitivity of temperature $\tau$ in terms of Micro-F1.}
    \label{fig:tau analysis}
\end{figure}

\subsubsection{Temperature}

The value of temperature $\tau$ determines the data distribution when measuring the distance between data points in contrasting. As illustrated in Figure \ref{fig:tau analysis}, we can see that our model is not sensitive to the temperature and have higher scores with lower variance against HeCo, showing its robustness. In addition, we observe that if the value of temperature is smaller, the gap between HGCML-P and HGCML will be larger. It is because the data distribution between positives and negatives will be smoother with the increase in temperature. The observation further proves the effectiveness of the proposed positive sampling strategy, especially with a small temperature.

\subsection{Visualization}

\begin{figure}[!t]
    \centering
    \subfigure[MP2vec]{
        \includegraphics[width=0.1\textwidth]{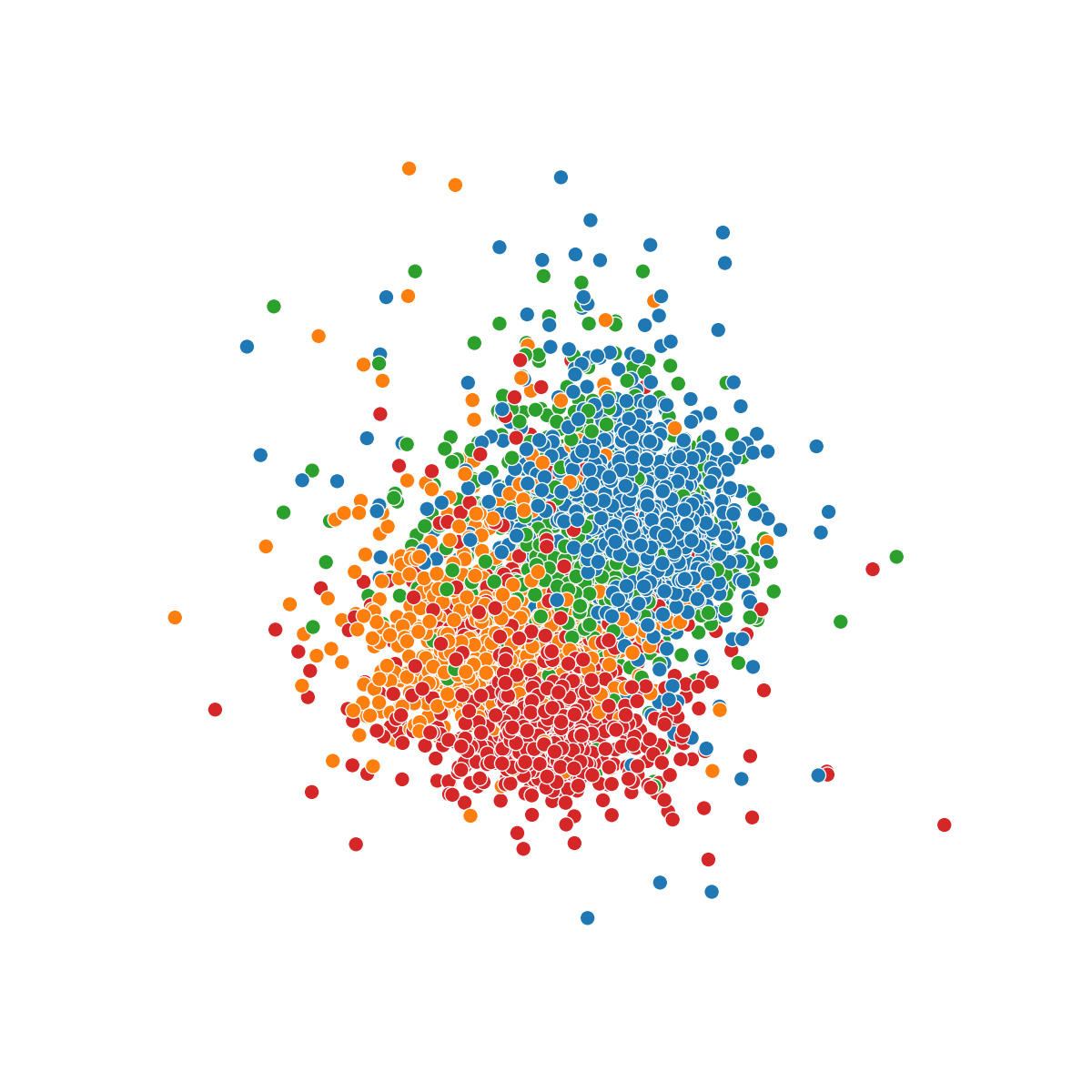}
        \label{fig:visual mp2vec}
    }
    \subfigure[DMGI]{
        \includegraphics[width=0.09\textwidth]{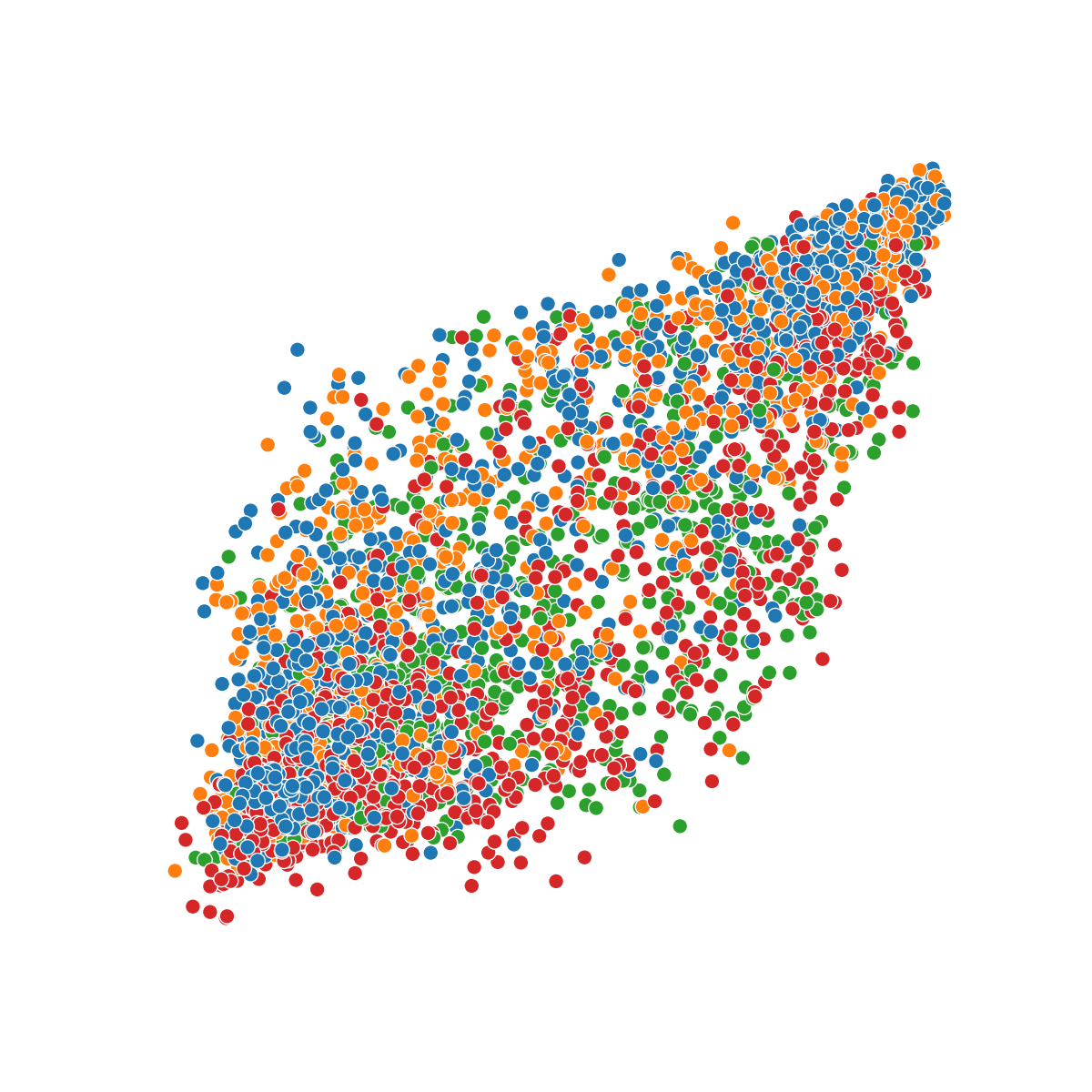}
        \label{fig:visual dmgi}
    }
    \subfigure[HeCo]{
        \includegraphics[width=0.1\textwidth]{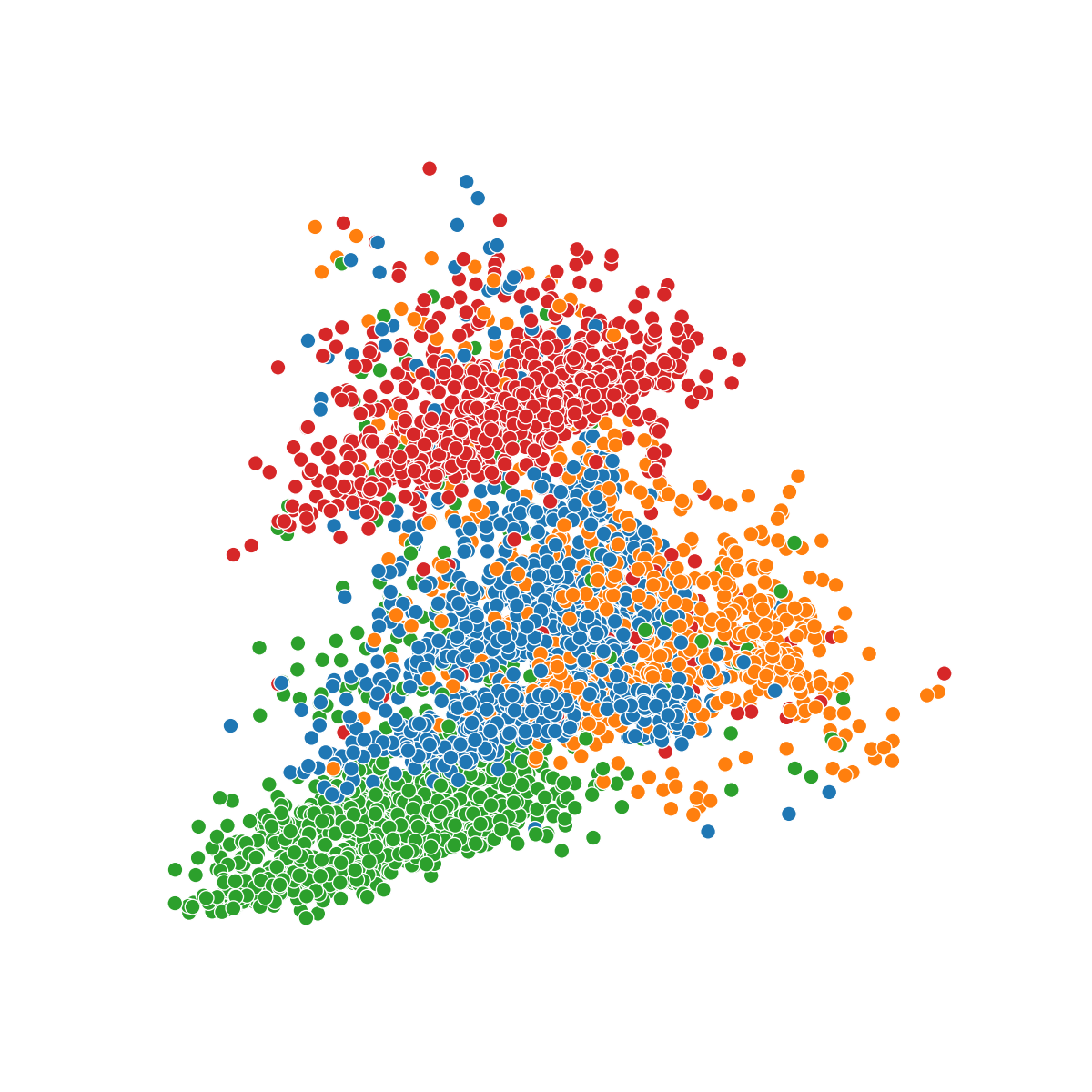}
        \label{fig:visual heco}
    }
    \subfigure[HGCML-P]{
        \includegraphics[width=0.1\textwidth]{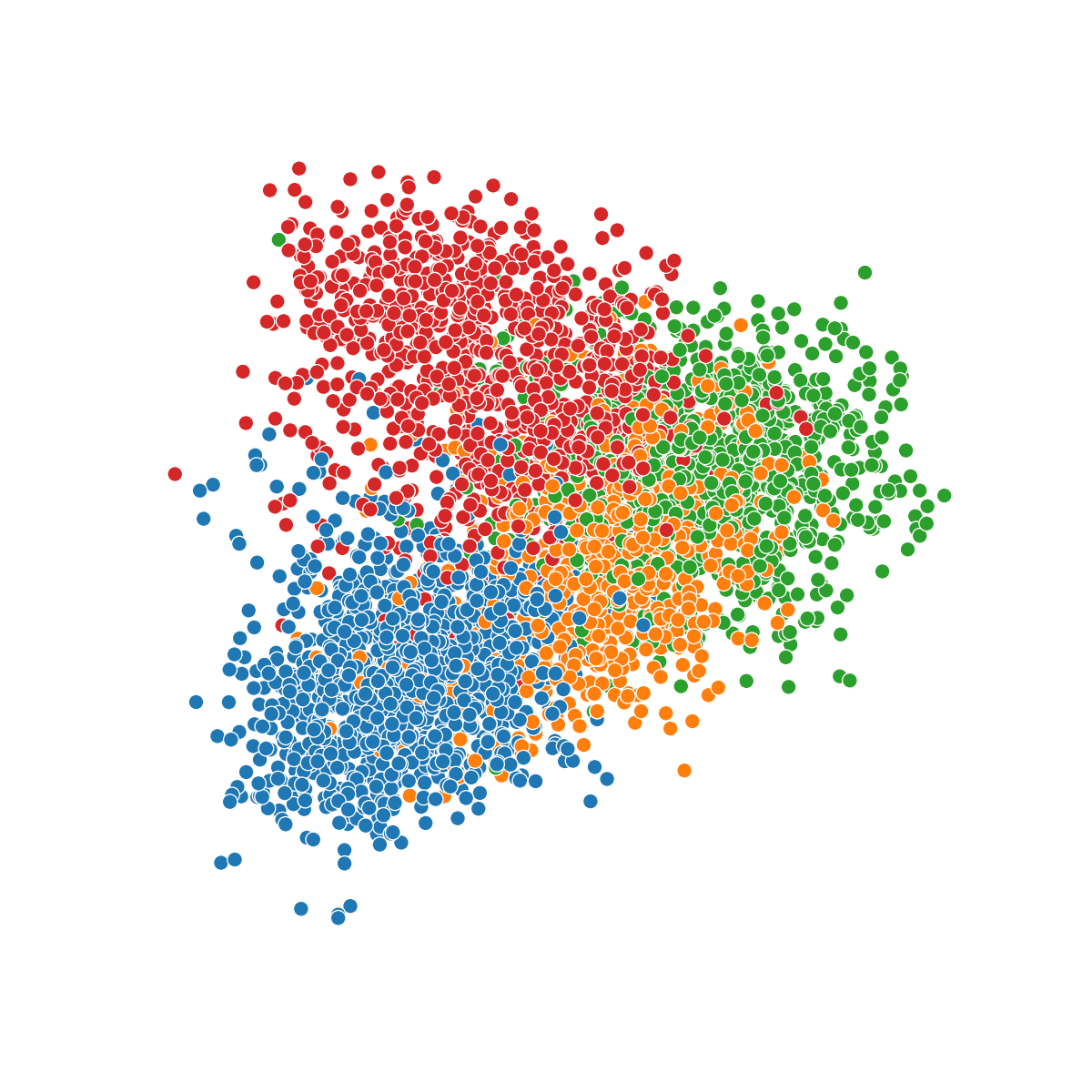}
        \label{fig:visual HGCML}
    }
    \caption{Visualization of node representations on DBLP.}
    \label{fig:visualization}
\end{figure}

To profoundly study the expressiveness of HGCML, we visualize the learned node representations of DBLP through $t$-SNE. In Figure \ref{fig:visualization}, we visualize node representations obtained from four algorithms, including Metapath2vec (MP2vec), DMGI, HeCo, and HGCML. As we can see, DMGI presents blurred boundaries between different classes, failing to learn discriminative low-dimensional node representations. For Metapath2vec and HeCo, despite some types of nodes being categorized clearly, there still exists a large proportion of overlapped data points that cannot be clearly identified. Our model separates nodes into different types, achieving the best performance.

\section{Conclusion}

In this paper, we propose a heterogeneous graph contrastive multi-view learning framework named HGCML. By treating metapaths as data augmentation, we create multi-views without impairing the underlying semantics in HINs. Then, we propose a novel objective that jointly performs intra-metapath and inter-metapath contrasts to model the consistency between metapaths. Specifically, we iteratively utilize graph patches and graph summaries to generate supervision signals to acquire local and global knowledge. To further enhance the quality of representations, we employ a positive sampling strategy that simultaneously considers node attributes and centrality to explicitly select positive samples to mitigate the sampling bias. Experimental results demonstrate the superiority of HGCML across five real-world datasets on node classification and node clustering.

\section{Acknowledgement}

This work was supported in part by the Zhejiang Provincial Natural Science Foundation of China under Grant No. LZ22F020002 and No. LY22F020003, and the National Natural Science Foundation of China under Grant No. 62002226 and No. 62002227.

\bibliographystyle{IEEETrans}
\bibliography{hgcml}

\end{document}